\documentclass[10pt,onecolumn]{article}

\usepackage{times}
\usepackage[dvips]{graphicx}
\usepackage[hang,scriptsize]{subfigure}
\usepackage{latexsym}

\usepackage{amsmath} % Extra environments for multiline displayed eq
\usepackage{amssymb} % Defines symbol names 
\usepackage{graphics} % nŽcessaire avec le package listings
\makeatletter
\def\@maketitle{%
  \vbox to 6.5cm{%
    \hsize\textwidth
    \linewidth\hsize
    \vspace{1.5cm}
    \centering
    {\bfseries\LARGE \@title \par}
    \vspace{12pt}
    {\fontsize{11pt}{13pt}\selectfont \begin{tabular}[t]{c}\@author \end{tabular}\par}
    \vfill} 
}
\renewcommand\section{\@startsection{section}{1}{\z@}%
                       {-12\p@ \@plus -4\p@ \@minus -4\p@}%
                       {6\p@ \@plus 4\p@ \@minus 4\p@}%
                       {\normalfont\large\bfseries
                        \rightskip=\z@ \@plus 8em\pretolerance=10000 }}
\renewcommand\subsection{\@startsection{subsection}{2}{\z@}%
                       {-12\p@ \@plus -4\p@ \@minus -4\p@}%
                       {6\p@ \@plus 4\p@ \@minus 4\p@}%
                       {\normalfont\fontsize{11pt}{13pt}\selectfont\bfseries
                        \rightskip=\z@ \@plus 8em\pretolerance=10000 }}
\renewcommand\subsubsection{\@startsection{subsubsection}{3}{\z@}%
                       {-12\p@ \@plus -4\p@ \@minus -4\p@}%
                       {6\p@ \@plus 4\p@ \@minus 4\p@}%
                       {\normalfont\normalsize\itshape}}
\renewcommand\paragraph{\@startsection{paragraph}{4}{\z@}%
                       {-12\p@ \@plus -4\p@ \@minus -4\p@}%
                       {-0.5em \@plus -0.22em \@minus -0.1em}%
                       {\normalfont\normalsize\itshape}}
\makeatother

\renewenvironment{abstract}%
  {\small
    \list{}{\labelwidth0pt
      \leftmargin0pt \rightmargin\leftmargin
      \listparindent\parindent \itemindent0pt
      \parsep0pt
      }%
    \item[\hskip\labelsep\bfseries\abstractname\enspace --] \itshape}{\endlist}

\newcommand{\keywordsname}{Keywords}
\newenvironment{keywords}%
  {\small
    \list{}{\labelwidth0pt
      \leftmargin0pt \rightmargin\leftmargin
      \listparindent\parindent \itemindent0pt
      \parsep0pt
      }%
    \item[\hskip\labelsep\bfseries\keywordsname:]}{\endlist}

\begin{document}

\title{The Generalized Pignistic Transformation}

% For three addresses
\author{\begin{tabular}{c@{\extracolsep{1em}}c@{\extracolsep{1em}}c}
{\bf Jean Dezert} & {\bf Florentin Smarandache} & {\bf Milan Daniel}\thanks {Partial support by the COST action 274 TARSKI  acknowledged.}\\
ONERA & Dpt.of Mathematics & Institute of Computer Science\\
9 Av. de la  Div. Leclerc & Univ. of New Mexico & Academy of Sciences of the Czech Republic\\
92320 Ch\^{a}tillon & Gallup, NM 8730 & Pod vod\'{a}renskou v\v{e}\v{z}\'{\i} 2,\ CZ - 182 07 \ Prague 8\\
France & U.S.A. & Czech Republic\\
{\tt Jean.Dezert@onera.fr} & {\tt smarand@unm.edu} & {\tt milan.daniel@cs.cas.cz}
\end{tabular}}

\date{}
\maketitle
\thispagestyle{empty}
\pagestyle{empty}

%%%%%%%%%%%%%%%%%%%%%%%%%%%%%%%%%%
% Text body

\begin{abstract}
This paper presents in detail the generalized pignistic transformation (GPT) succinctly developed in the Dezert-Smarandache Theory (DSmT) framework as a tool for decision process. The GPT allows to provide a subjective probability measure from any generalized basic belief assignment given by any corpus of evidence. We mainly focus our presentation on the 3D case and provide the complete result obtained by the GPT  and its validation drawn from the probability theory. 
\end{abstract}


%\noindent {\bf{MSC 2000}}: 68T37, 94A15, 94A17, 68T40.

\section{Introduction}
%********************

In the recent theory of plausible and paradoxical reasoning (DSmT) developed by Dezert and Smarandache \cite{Dezert_2003,Smarandache_2002}, a new generalized pignistic transformation has been proposed to construct a subjective probability measure $P\{.\}$ from any generalized basic belief assignment $m(.)$ defined over the hyper-power set $D^\Theta$. In reference \cite{Dezert_2003}, a simple example of such generalized pignistic transformation has been presented only for the case $n=|\Theta|=2$. In this paper, we present the complete derivation of this pignistic transformation for the case $n=|\Theta|=3$ and we generalize the result. Before introducing the GPT, it is however necessary to briefly present the DSmT \cite{Dezert_2002b, Dezert_2003,Dezert_Smarandache_2003,Dezert_2003f,Dezert_2004,Smarandache_2002} with respect to the Dempster-Shafer Theory (DST) \cite{Shafer_1976}.

\section{Foundations of the DST and DSmT}
%**********************************************

\subsection{The DST and the Shafer's model}

The Shafer's model, denoted here $\mathcal{M}^0(\Theta)$, on which is based the Dempster-Shafer Theory, assumes an exhaustive and exclusive frame of discernment of the problem under consideration $\Theta=\{\theta_1,\theta_2,\ldots,\theta_n\}$.  The model requires actually that an ultimate refinement of the problem is possible so that $\theta_i$ can always be well precisely defined/identified in such a way  that we are sure that they are exclusive and exhaustive. From this model,  a basic belief assignment (bba) $m_i(.): 2^\Theta \rightarrow  [0, 1]$  such that $m_i(\emptyset)=0$ and $\sum_{A\in 2^\Theta} m_i(A) = 1$ associated to a given body of evidence $\mathcal{B}_i$ is defined, where $2^\Theta$ is the {\it{power set}} of $\Theta$, i.e. the set of all subsets of $\Theta$. Within DST, the fusion (combination) of two independent sources of evidence $\mathcal{B}_1$ and $\mathcal{B}_2$ is obtained through the Dempster's rule of combination  \cite{Shafer_1976} : $[m_{1}\oplus m_{2}](\emptyset)=0$ and $\forall B\neq\emptyset \in 2^\Theta$:
 \begin{equation}
 [m_{1}\oplus m_{2}](B) = 
\frac{\sum_{X\cap Y=B}m_{1}(X)m_{2}(Y)}{1-\sum_{X\cap Y=\emptyset} m_{1}(X) m_{2}(Y)} 
\label{eq:DSR}
 \end{equation}
 
The notation $\sum_{X\cap Y=B}$ represents the sum over all $X, Y \in 2^\Theta$ such that $X\cap Y=B$. The Dempster's sum $m (.)\triangleq [m_{1}\oplus m_{2}](.)$ is considered as a basic belief assignment if and only if the denominator in equation \eqref{eq:DSR} is non-zero. The term $k_{12}\triangleq \sum_{X\cap Y=\emptyset} m_{1}(X) m_{2}(Y)$ is called degree of conflict between the sources $\mathcal{B}_1$ and $\mathcal{B}_2$. When $k_{12}=1$,  the Dempster's sum $m (.)$ does not exist and the bodies of evidences $\mathcal{B}_1$ and $\mathcal{B}_2$ are said to be in {\it{full contradiction}}. This rule of combination can be extended easily for the combination of $n>2$ independent sources of evidence. The DST, although very attractive because of its solid mathematical ground,  presents however several weaknesses and limitations because of the Shafer's model itself (which does not necessary hold in some fusion problems involving continuous and ill-defined concepts), the justification of the Dempster's rule of combination frequently subject to criticisms, but mainly because of counter-intuitive results given by the Dempster's rule when the conflict between sources becomes important. Several classes of infinite counter-examples to the Dempster's rule can be found in \cite{Dezert_2004K}. To overcome these limitations, Jean Dezert and Florentin Smarandache propose a new mathematical theory  based on other models (free or hybrid DSm models) with new reliable rules of combinations able to deal with any kind of sources ( imprecises, uncertain and paradoxist, i.e. highly conflicting). This is presented in next subsections.

\subsection{The DSmT based on the free DSm Model}
%*********************************************************

The foundations of the DSmT (Dezert-Smarandache Theory) is to abandon the Shafer's model (i.e. the exclusivity constraint between $\theta_i$ of $\Theta$) just because for some fusion problems it is impossible to define/characterize the problem in terms of well-defined/precise and exclusive elements.  The free DSm model, denoted $\mathcal{M}^f(\Theta)$, on which is based DSmT allows us to deal with  imprecise/vague notions and concepts between elements of the frame of discernment $\Theta$. The DSmT  includes the possibility to deal with evidences arising from different sources of information which don't have 
access to absolute interpretation of the elements $\Theta$ under consideration.

%***************************************************
\subsubsection{Notion of hyper-power set $D^\Theta$}
%***************************************************

From this very simple idea and from any frame $\Theta$, a new space $D^\Theta=\{\alpha_0,\ldots,\alpha_{d(n)-1}\}$ (free Boolean pre-algebra generated by $\Theta$ and operators $\cap$ and $\cup$), called {\it{hyper-power set}} is defined \cite{Dezert_2004} as follows:
\begin{enumerate}
\item $\emptyset, \theta_1,\ldots, \theta_n \in D^\Theta$
\item $\forall A\in D^\Theta, B\in D^\Theta, (A\cup B)\in D^\Theta, (A\cap B)\in D^\Theta$
\item No other elements belong to $D^\Theta$, except those, obtained by using rules 1 or 2.
\end{enumerate}
The generation of hyper-power set $D^\Theta$ is related with the famous Dedekind's problem on enumerating the set of monotone Boolean functions. The cardinality $d(n)$ of $D^\Theta$ follows the Dedekind sequence. It can be shown, see \cite{Dezert_2003f}, that all elements $\alpha_i$ of $D^\Theta$ can then be obtained by the very simple linear equation \cite{Dezert_2003f}
\begin{equation}
\mathbf{d}_n= \mathbf{D}_n\cdot\mathbf{u}_n
\end{equation}
\noindent where $\mathbf{d}_n\equiv[\alpha_0\equiv\emptyset,\alpha_1,\ldots,\alpha_{d(n)-1}]'$ is the vector of elements of $D^\Theta$, $\mathbf{u}_n$ is the proper Smarandache's codification vector \cite{Dezert_2003f} and $D_n$ a particular binary matrix build recursively by the algorithm proposed in \cite{Dezert_2003f}. The final result $\mathbf{d}_n$ is obtained from the previous {\it{matrix product}} after identifying $(+,\cdot)$ with $(\cup,\cap)$ operators, $0\cdot x$ with $\emptyset$ and $1\cdot x$ with $x$). $D_n$ is actually a binary matrix corresponding to all possible isotone Boolean functions.

%***************************************************
\subsubsection{Classic DSm rule of combination}
%***************************************************

By adopting the free DSm model and from any general frame of discernment $\Theta$, one then defines a map $m_i(.): D^\Theta \rightarrow [0,1]$, associated to a given source of evidence $\mathcal{B}_i$ such that $m_i(\emptyset)=0$ and $\sum_{A\in D^\Theta} m_i(A) = 1$. This approach allows us to model any source which supports paradoxical (or intrinsic conflicting) information. From this very simple free DSm model $\mathcal{M}^f(\Theta)$, the classical DSm rule of combination $m(.)\triangleq [m_{1}\oplus \ldots \oplus m_{k}](.)$ of $k\geq 2$ intrinsic conflicting and/or uncertain independent sources of information is defined by \cite{Dezert_2002b}
\begin{equation}
m_{\mathcal{M}^f(\Theta)}(A)=
\sum_{\overset{X_1,\ldots,X_k\in D^\Theta}{X_1\cap \ldots\cap X_k=A}} \prod_{i=1}^{k}m_i(X_i)
\label{eq:DSMClassick}
\end{equation}
\noindent
and $m_{\mathcal{M}^f(\Theta)}(\emptyset)=0$ by definition. This rule, dealing with uncertain and/or paradoxical/conflicting information is commutative and associative and requires no normalization procedure.

\subsection{Extension of the DSmT to hybrid models}
%*********************************************************

\subsubsection{Notion of hybrid model}

The adoption of the free DSm model (and the classic DSm rule) versus the Shafer's model (with the Dempster's rule) can also be subject to criticisms since not all fusion problems correspond to the free DSm model (neither to the Shafer's model). These two models can be viewed actually as  the two opposite/extreme and specific models on which are based the DSmT and the DST. In general, the models for characterizing practical fusion problems do not coincide neither with the Shafer's model nor with the free DSm model. They have an hybrid nature (only some $\theta_i$ are truly exclusive).Very recently, F. Smarandache and J. Dezert have extended the framework of the DSmT and the previous DSm rule of combination for solving a wider class of fusion problems in which neither free DSm or Shafer's models fully hold. This large class of problems corresponds to problems characterized by any hybrid DSm model. A hybrid DSm model is defined from the free DSm model $\mathcal{M}^f(\Theta)$ by introducing some integrity constraints on some elements $A\in D^\Theta$, if there are some certain facts in accordance with the exact nature of the model related to the problem under consideration \cite{Dezert_2004b}. An integrity constraint on $A\in D^\Theta$ consists in forcing $A$ to be empty through the model $\mathcal{M}$, denoted as $A\overset{\mathcal{M}}{\equiv}\emptyset$. There are several possible kinds of integrity constraints introduced in any free DSm model:
\begin{itemize}
\item {\it{Exclusivity constraints}}: when some conjunctions of elements of $\Theta$
 are truly impossible, for example when  $\theta_i\cap\ldots\cap\theta_k\overset{\mathcal{M}}{\equiv}\emptyset$.
\item {\it{Non-existential constraints}}: when some disjunctions of elements of $\Theta$ are truly impossible, for example when 
$\theta_i\cup\ldots\cup\theta_k\overset{\mathcal{M}}{\equiv}\emptyset$. The degenerated hybrid DSm model $\mathcal{M}_{\emptyset}$, 
defined by constraint according to the total ignorance: $I_t\triangleq\theta_1\cup \theta_2\cup \ldots\cup \theta_n\overset{\mathcal{M}}{\equiv}\emptyset$, is excluded from consideration, because it is meaningless.
\item {\it{Hybrid constraints}}: like for example $(\theta_i\cap\theta_j)\cup \theta_k\overset{\mathcal{M}}{\equiv}\emptyset$ and any other hybrid proposition/element of $D^\Theta$ involving both $\cap$ and $\cup$ operators such that at least one element $\theta_k$ is subset of the constrained proposition.
\end{itemize}
The introduction of a given integrity constraint $A\overset{\mathcal{M}}{\equiv}\emptyset \in D^\Theta$ implies the set of inner constraints $B\overset{\mathcal{M}}{\equiv}\emptyset$ for all $B\subset A$. The introduction of two integrity constraints on $A,B \in D^\Theta$ implies the constraint $(A\cup B)\in D^\Theta \equiv\emptyset$ and this implies the emptiness of all $C\in D^\Theta$ such that $C\subset (A\cup B)$.\\

The Shafer's model, denoted $\mathcal{M}^{0}(\Theta)$, can be considered as the most constrained hybrid DSm model including all possible exclusivity constraints {\it{without non-existential constraint}}, since all elements in the frame are forced to be mutually exclusive. 

\subsubsection{The hybrid DSm rule of combination}

The hybrid DSm rule of combination, associated to a given hybrid DSm model $\mathcal{M}\neq\mathcal{M}_{\emptyset}$ , for $k\geq 2$ independent sources of information is defined for all $A\in D^\Theta$ as \cite{Dezert_2004b}:
\begin{equation}
m_{\mathcal{M}(\Theta)}(A)\triangleq 
\phi(A)\Bigl[ S_1(A) + S_2(A) + S_3(A)\Bigr]
 \label{eq:DSmHkBis}
\end{equation}
\noindent
where $\phi(A)$ is the characteristic non emptiness function of a set $A$, i.e. $\phi(A)= 1$ if  $A\notin \boldsymbol{\emptyset}$ and $\phi(A)= 0$ otherwise, where $\boldsymbol{\emptyset}\triangleq\{\boldsymbol{\emptyset}_{\mathcal{M}},\emptyset\}$. $\boldsymbol{\emptyset}_{\mathcal{M}}$ is the set  of all elements of $D^\Theta$ which have been forced to be empty through the constraints of the model $\mathcal{M}$ and $\emptyset$ is the classical/universal empty set. $S_1(A)\equiv m_{\mathcal{M}^f(\Theta)}(A)$, $S_2(A)$, $S_3(A)$ are defined by \cite{Dezert_2004b}
\begin{equation}
S_1(A)\triangleq \sum_{\substack{X_1,X_2,\ldots,X_k\in D^\Theta \\ (X_1\cap X_2\cap\ldots\cap X_k)=A}} \prod_{i=1}^{k} m_i(X_i)
\end{equation}
\begin{equation}
S_2(A)\triangleq \sum_{\substack{X_1,X_2,\ldots,X_k\in\boldsymbol{\emptyset} \\  [\mathcal{U}=A]\vee [(\mathcal{U}\in\boldsymbol{\emptyset}) \wedge (A=I_t)]}} \prod_{i=1}^{k} m_i(X_i)\end{equation}
\begin{equation}
S_3(A)\triangleq\sum_{\substack{X_1,X_2,\ldots,X_k\in D^\Theta \\ (X_1\cup X_2\cup\ldots\cup X_k)=A \\ (X_1\cap X_2\cap \ldots\cap X_k)\in\boldsymbol{\emptyset}}}  \prod_{i=1}^{k} m_i(X_i)
\end{equation}
with $\mathcal{U}\triangleq u(X_1)\cup u(X_2)\cup \ldots \cup u(X_k)$ where $u(X)$ is the union of all singletons $\theta_i$ that compose $X$ and $I_t \triangleq \theta_1\cup \theta_2\cup\ldots\cup \theta_n$ is the total ignorance. $S_1(A)$ corresponds to the classic DSm rule of combination based on the free DSm model; $S_2(A)$ represents the mass of all relatively and absolutely empty sets which is transferred to the total or relative ignorances; $S_3(A)$ transfers the sum of relatively empty sets to the non-empty sets.

\subsection{The DSm cardinality $\mathcal{C}_\mathcal{M}(A)$}
%**************************************

\subsubsection{Definition}

One important notion involved in the definition of the generalized pignistic transformation (GPT) is the {\it{DSm cardinality}} \cite{Dezert_Smarandache_2003}. The {\it{DSm cardinality}} of any element $A \in D^\Theta$, denoted $\mathcal{C}_\mathcal{M}(A)$, corresponds to the number of parts of $A$ in the Venn diagram of the problem (model $\mathcal{M}$) taking into account the set of integrity constraints (if any), i.e. all the possible intersections due to the nature of the elements $\theta_i$. This {\it{intrinsic cardinality}} depends on the model $\mathcal{M}$ (free, hybrid or Shafer's model).  $\mathcal{M}$ is the model that contains $A$, which depends both on the
dimension $n=\vert \Theta \vert$ and on the number of parts of non-empty intersections present in its associated Venn diagram.
%
%Venn diagram, (i.e. the number of sets $n=\vert \Theta \vert$ under consideration),
%and on the number of non-empty intersections in this diagram. 
One has $1 \leq \mathcal{C}_\mathcal{M}(A) \leq 2^n-1$. $\mathcal{C}_\mathcal{M}(A)$ must not be confused with the classical cardinality $\vert A \vert$ of a given set $A$ (i.e. the number of its distinct elements) - that's why a new notation is necessary here. \\

It can be shown, see \cite{Dezert_Smarandache_2003}, that $\mathcal{C}_\mathcal{M}(A)$, is exactly equal to the sum of the elements of the row of $\mathbf{D}_n$ corresponding to proposition $A$ in the $\mathbf{u}_n$ basis (see section 2.1.1). Actually $\mathcal{C}_\mathcal{M}(A)$ is very easy to compute by programming from the algorithm of generation of $D^\Theta$ given in \cite{Dezert_2003f}.\\

If one imposes a constraint that a set $B$ from $D^\Theta$ is empty (i.e. we choose a hybrid model), then one suppresses the columns corresponding to the parts which compose $B$ in the matrix $\mathbf{D}_n$ and the row of $B$ and the rows of all elements  of $D^\Theta$ which are subsets of $B$, getting a new matrix ${\mathbf{D}'}_n$ which represents a new hybrid model $\mathcal{M}'$. In the $\mathbf{u}_n$ basis, one similarly suppresses the parts that form $B$, and now this basis has the dimension $2^n-1-\mathcal{C}_{\mathcal{M}}(B)$.

\subsubsection{A 3D example with the free DSm model $\mathcal{M}^f$}

Consider the 3D case $\Theta=\{\theta_1,\theta_2,\theta_3\}$ with the free DSm model $\mathcal{M}^f$  corresponding to the following Venn diagram (where $<i>$ denotes the part 
which belongs to $\theta_i$ only, $<ij>$ denotes the part which belongs to $\theta_i$ and $\theta_j$ only, etc; this is the Smarandache's codification \cite{Dezert_2003f}).

\begin{center}
{\tt \setlength{\unitlength}{1pt}
\begin{picture}(90,90)
\thinlines    
\put(40,60){\circle{40}}
\put(60,60){\circle{40}}
\put(50,40){\circle{40}}
\put(15,84){\vector(1,-1){10}}
\put(7,84){$\theta_{1}$}
\put(84,84){\vector(-1,-1){10}}
\put(85,84){$\theta_{2}$}
\put(74,15){\vector(-1,1){10}}
\put(75,10){$\theta_{3}$}
\put(45,64){\tiny{<12>}}
\put(42,50){\tiny{<123>}}
\put(47,30){\tiny{<3>}}
\put(32,43){\tiny{<13>}}
\put(58,43){\tiny{<23>}}
\put(68,60){\tiny{<2>}}
\put(28,60){\tiny{<1>}}
\end{picture}}
\end{center}
The elements of $D^\Theta$ with their DSm cardinality are given by the following table:
\begin{equation*}
\begin{array}{lcl}
A\in D^\Theta                     & \mathcal{C}_{\mathcal{M}^f}(A) \\
\hline
\alpha_0\triangleq\emptyset                                                                   & 0 \\
\alpha_1\triangleq\theta_1\cap\theta_2\cap\theta_3       & 1 \\
\alpha_2\triangleq\theta_1\cap\theta_2                              & 2 \\
\alpha_3\triangleq\theta_1\cap\theta_3                              & 2 \\
\alpha_4\triangleq\theta_2\cap\theta_3                              & 2 \\
\alpha_5\triangleq(\theta_1\cup\theta_2)\cap\theta_3     & 3 \\
\alpha_6\triangleq(\theta_1\cup\theta_3)\cap\theta_2      & 3 \\
\alpha_7\triangleq(\theta_2\cup\theta_3)\cap\theta_1      & 3 \\
\alpha_8\triangleq\{(\theta_1\cap\theta_2)\cup\theta_3\} \cap(\theta_1\cup\theta_2) & 4 \\
\alpha_9\triangleq\theta_1                                                      & 4 \\
\alpha_{10}\triangleq\theta_2                                                      & 4 \\
\alpha_{11}\triangleq\theta_3                                                      & 4 \\
\alpha_{12}\triangleq(\theta_1\cap\theta_2)\cup\theta_3       &  5 \\
\alpha_{13}\triangleq(\theta_1\cap\theta_3)\cup\theta_2       & 5 \\
\alpha_{14}\triangleq(\theta_2\cap\theta_3)\cup\theta_1       & 5 \\
\alpha_{15}\triangleq\theta_1\cup\theta_2                             & 6 \\
\alpha_{16}\triangleq\theta_1\cup\theta_3                             & 6 \\
\alpha_{17}\triangleq\theta_2\cup\theta_3                             & 6 \\
\alpha_{18}\triangleq\theta_1\cup\theta_2\cup\theta_3         & 7  \\
\end{array}
\end{equation*}
\begin{center}
{\bf{Table 1}}: $\mathcal{C}_{\mathcal{M}^f}(A)$ for free DSm model $\mathcal{M}^f$
\end{center}

\subsubsection{A 3D example with a given hybrid model}

Consider now the same 3D case with the model $\mathcal{M}\neq\mathcal{M}^f$ in which we force all possible conjunctions to be empty, but $\theta_1\cap\theta_2$ according to the following Venn diagram.\\
\begin{center}
{\tt \setlength{\unitlength}{1pt}
\begin{picture}(90,90)
\thinlines    
\put(40,60){\circle{40}}
\put(60,60){\circle{40}}
\put(50,10){\circle{40}}
\put(15,84){\vector(1,-1){10}}
\put(7,84){$\theta_{1}$}
\put(84,84){\vector(-1,-1){10}}
\put(85,84){$\theta_{2}$}
\put(85,10){\vector(-1,0){15}}
\put(87,7){$\theta_{3}$}
\put(45,59){\tiny{<12>}}
\put(47,7){\tiny{<3>}}
\put(68,59){\tiny{<2>}}
\put(28,59){\tiny{<1>}}
\end{picture}}
\end{center}

Then, one gets the following list of elements (with their DSm cardinal) for the restricted $D^\Theta$ taking into account the integrity constraints of this hybrid model:
\begin{equation*}
\begin{array}{lcl}
A\in D^\Theta                     & \mathcal{C}_{\mathcal{M}}(A) \\
\hline
\alpha_0\triangleq\emptyset                                                 & 0 \\
\alpha_1\triangleq\theta_1\cap\theta_2                             & 1 \\
\alpha_2\triangleq\theta_3                                                    & 1 \\
\alpha_3\triangleq\theta_1                                                    & 2 \\
\alpha_4\triangleq\theta_2                                                    & 2 \\
\alpha_5\triangleq\theta_1\cup\theta_2                             & 3 \\
\alpha_6\triangleq\theta_1\cup\theta_3                             & 3 \\
\alpha_7\triangleq\theta_2\cup\theta_3                             & 3 \\
\alpha_8\triangleq\theta_1\cup\theta_2\cup\theta_3       & 4  \\
\end{array}
\end{equation*}
\begin{center}
{\bf{Table 2}}: $\mathcal{C}_{\mathcal{M}}(A)$ for the chosen hybrid model $\mathcal{M}$
\end{center}

\subsubsection{A 3D example with the Shafer's model}
Consider now the same 3D case but with all exclusivity constraints on $\theta_i$, $i=1,2,3$. This corresponds to the 3D Shafer's model $\mathcal{M}^0$ presented in the following Venn diagram.\\
\begin{center}
{\tt \setlength{\unitlength}{1pt}
\begin{picture}(90,90)
\thinlines    
\put(20,60){\circle{40}} % cercle theta 1
\put(80,60){\circle{40}} % cercle theta 2
\put(50,10){\circle{40}} % cercle theta 3
\put(-5,84){\vector(1,-1){10}}
\put(-13,84){$\theta_{1}$}
\put(104,84){\vector(-1,-1){10}}
\put(105,84){$\theta_{2}$}
\put(85,10){\vector(-1,0){15}}
\put(87,7){$\theta_{3}$}
\put(47,7){\tiny{<3>}}
\put(75,59){\tiny{<2>}}
\put(15,59){\tiny{<1>}}
\end{picture}}
\end{center}

Then, one gets the following list of elements (with their DSm cardinal) for the restricted $D^\Theta$, which coincides naturally with the classical power set $2^\Theta$:
\begin{equation*}
\begin{array}{lcl}
A\in (D^\Theta\equiv 2^\Theta)                     & \mathcal{C}_{\mathcal{M}^0}(A) \\
\hline
\alpha_0\triangleq\emptyset                                                 & 0 \\
\alpha_1\triangleq\theta_1                                                    & 1 \\
\alpha_2\triangleq\theta_2                                                    & 1 \\
\alpha_3\triangleq\theta_3                                                    & 1 \\
\alpha_4\triangleq\theta_1\cup\theta_2                             & 2 \\
\alpha_5\triangleq\theta_1\cup\theta_3                             & 2 \\
\alpha_6\triangleq\theta_2\cup\theta_3                             & 2 \\
\alpha_7\triangleq\theta_1\cup\theta_2\cup\theta_3       & 3  \\
\end{array}
\end{equation*}
\begin{center}
{\bf{Table 3}}: $\mathcal{C}_{\mathcal{M}}(A)$ for the 3D Shafer's model $\mathcal{M}^0$
\end{center}

%**********************************
\section{ The pignistic transformations}
%**********************************

We follow here the Smets' point of view \cite{Smets_2000} about the assumption that beliefs manifest themselves at two mental levels: the {\it{credal}} level where beliefs are entertained and the {\it{pignistic}} level where belief are used to make decisions. Pignistic terminology has been coined by Philippe Smets and comes from {\it{pignus}}, a bet in Latin.  The probability functions, usually used to quantify beliefs at both levels, are actually used here only to quantify the uncertainty when a decision is really necessary, otherwise we argue as Philippe Smets does, that beliefs are represented by belief functions. To take a rational decision, we propose to transform beliefs into pignistic probability functions through the generalized pignistic transformation (GPT) which will be presented in the sequel. We first recall the classical pignistic transformation (PT) based on the DST and then we generalize it within the DSmT framework.

\subsection{The classical pignistic transformation}
%*************************************************

When a decision must be taken, we use the expected utility theory which requires to construct a probability function $P\{.\}$ from basic belief assignment $m(.)$ \cite{Smets_2000}. This is achieved by the so-called classical pignistic transformation\footnote{We don't divide here $m(X)$ by $1-m(\emptyset)$ as in the P. Smets' formulation just because $m(\emptyset)=0$ in the DSmT framework, unless there is a solid necessity to justify to do it.} as follows (see \cite{Smets_1994} for justification):

\begin{equation}
P\{A\}=\sum_{X \in 2^\Theta}\frac{|X\cap A|}{|X|}m(X)
\label{eq:Pig}
\end{equation}

\noindent
where $|A|$ denotes the number of worlds in the set $A$ (with convention $|\emptyset | / |\emptyset |=1$, to define $P\{\emptyset \}$). $P\{A\}$ corresponds to $BetP(A)$ in the Smets' notation \cite{Smets_2000}. Decisions are achieved by computing the expected utilities of the acts using the subjective/pignistic $P\{.\}$ as the probability function needed to compute expectations.
Usually, one uses the maximum of the pignistic probability as decision criterion. The max. of $P\{.\}$ is often considered as a prudent betting decision criterion between the two other alternatives (max of plausibility or max. of credibility). It is easy to show that $P\{.\}$ is indeed a probability function (see \cite{Smets_1994}).
% which appears to be respectively too optimistic or too pessimistic). 

\subsection{The generalized pignistic transformation}
%------------------------------------------------------------------

\subsubsection{Definition}

To take a rational decision within the DSmT framework, it is then necessary to generalize the classical pignistic transformation in order to construct a pignistic probability function from any generalized basic belief assignment $m(.)$ drawn form the DSm rule of combination (the classic or hybrid rule). This generalized pignistic transformation (GPT) is defined by: $\forall A \in D^\Theta$,
\begin{equation}
P\{A\}=\sum_{X \in D^\Theta}  \frac{\mathcal{C}_{\mathcal{M}}(X\cap A)}{\mathcal{C}_{\mathcal{M}}(X)}m(X)
\label{eq:PigG}
\end{equation}

\noindent
where $\mathcal{C}_{\mathcal{M}}(X)$ denotes the DSm cardinal of proposition $X$ for the DSm model $\mathcal{M}$ of the problem under consideration.\\

The decision about the solution of the problem  is usually taken by the maximum of pignistic probability function $P\{.\}$. Let's remark the close ressemblance of the two pignistic transformations \eqref{eq:Pig} and \eqref{eq:PigG}. It can be shown that \eqref{eq:PigG} reduces to \eqref{eq:Pig} when the hyper-power set $D^\Theta$ reduces to classical power set $2^\Theta$ if we adopt the Shafer's model. But  \eqref{eq:PigG} is a generalization of  \eqref{eq:Pig} since it can be used for computing pignistic probabilities for any models (including the Shafer's model).

\subsubsection{$P\{.\}$ is a probability measure}
%****************************************************************

It is important to prove that $P\{.\}$ built from GPT is indeed a (subjective/pignistic) probability measure satisfying the following axioms of the probability theory \cite{Li_1999,Papoulis_1989}:

\begin{itemize}
\item {\bf{Axiom 1}} (nonnegativity): The (generalized pignistic) probability of any event $A$ is bounded by 0 and 1, i.e. $0 \leq P\{A\} \leq 1$
\item {\bf{Axiom 2}} (unity): Any sure event (the sample space) has unity (generalized pignistic) probability, i.e. $P\{\mathcal{S}\}=1$
\item {\bf{Axiom 3}} (additivity over mutually exclusive events): If $A$, $B$ are disjoint (i.e. $A\cap B=\emptyset$) then $P(A\cup B)=P(A)+P(B)$
\end{itemize}

The axiom 1 is satisfied because, by the definition of the generalized basic belief assignment $m(.)$, one has $\forall \alpha_i\in D^\Theta$, $0 \leq m(\alpha_i) \leq 1$ with $\sum_{\alpha_i\in D^\Theta}m(\alpha_i)=1$ and since all coefficients involved within GPT are bounded by 0 and 1, it follows directly that pignistic probabilities are also bounded by 0 and 1.\\

 The axiom 2 is satisfied because all the coefficients involved in the sure event $\mathcal{S}\triangleq \theta_1\cup\theta_2\cup ...\cup \theta_n$ are equal to one because $\mathcal{C}_{\mathcal{M}}(X\cap\mathcal{S})/\mathcal{C}_{\mathcal{M}}(X)=\mathcal{C}_{\mathcal{M}}(X)/\mathcal{C}_{\mathcal{M}}(X)=1$, so that $P\{\mathcal{S}\}\equiv\sum_{\alpha_i\in D^\Theta}m(\alpha_i)=1$.\\
 
The axiom 3 is satisfied. Indeed, from the definition of GPT, one has
\begin{equation}
 P\{A\cup B\}=\sum_{X \in D^\Theta}  \frac{\mathcal{C}_{\mathcal{M}}(X\cap (A\cup B))}{\mathcal{C}_{\mathcal{M}}(X)}m(X)
 \label{eq:A3}
\end{equation}

But if we consider $A$ and $B$ exclusive (i.e. $A\cap B=\emptyset$), then it follows:
\begin{align*}
\mathcal{C}_{\mathcal{M}}(X\cap(A\cup B)) & =\mathcal{C}_{\mathcal{M}}((X\cap A)\cup (X\cap B))\\
& =\mathcal{C}_{\mathcal{M}}(X\cap A)+\mathcal{C}_{\mathcal{M}}(X\cap B)
\end{align*}

By substituting $\mathcal{C}_{\mathcal{M}}(X\cap(A\cup B))$ by $\mathcal{C}_{\mathcal{M}}(X\cap A)+\mathcal{C}_{\mathcal{M}}(X\cap B)$ into \eqref{eq:A3}, it comes:
\begin{align*}
 P\{A\cup B\} & =\sum_{X \in D^\Theta}  \frac{\mathcal{C}_{\mathcal{M}}(X\cap A)+\mathcal{C}_{\mathcal{M}}(X\cap B)}{\mathcal{C}_{\mathcal{M}}(X)}m(X)\\
& =\sum_{X \in D^\Theta} \frac{\mathcal{C}_{\mathcal{M}}(X\cap A)}{\mathcal{C}_{\mathcal{M}}(X)}m(X)\\
& \qquad \qquad 
 + \sum_{X \in D^\Theta} \frac{\mathcal{C}_{\mathcal{M}}(X\cap B)}{\mathcal{C}_{\mathcal{M}}(X)}m(X)\\
& = P\{A\} + P\{B\}
\end{align*}
\noindent
which completes the proof. From the coefficients $\frac{\mathcal{C}_{\mathcal{M}}(X\cap A)}{\mathcal{C}_{\mathcal{M}}(X)}$ involved in \eqref{eq:PigG}, it can also be easily checked that $A\subset B \Rightarrow P\{A\} \leq P\{B\}$. One can also easily prove the Poincar\'{e}' equality:
$P\{A\cup B\}=P\{A\}+P\{B\}-P\{A\cap B\}$ because $\mathcal{C}_{\mathcal{M}}(X\cap(A\cup B)= \mathcal{C}_{\mathcal{M}}((X\cap A)\cup(X\cap B))=\mathcal{C}_{\mathcal{M}}(X\cap A) + \mathcal{C}_{\mathcal{M}}(X\cap B)-\mathcal{C}_{\mathcal{M}}(X\cap (A\cap B))$
(one has substracted $\mathcal{C}_{\mathcal{M}}(X\cap (A\cap B))$, i.e. the number of parts of $X\cap (A\cap B)$ in the Venn diagram, due to the fact that these parts were added twice: once in $\mathcal{C}_{\mathcal{M}}(X\cap A)$ and second time in $\mathcal{C}_{\mathcal{M}}(X\cap B)$.

%***************************
\section{Examples of GPT}
%***************************

\subsection{Example for the 2D case}
%***************************************

\begin{itemize}
\item
{\bf{With the free DSm model}}: Let's  consider $\Theta=\{\theta_1,\theta_2\}$ and the generalized basic belief assignment $m(.)$ over the hyper-power set $D^\Theta=\{\emptyset,\theta_1\cap\theta_2,\theta_1,\theta_2,\theta_1\cup\theta_2\}$.
It is easy to construct the pignistic probability $P\{.\}$. According to the definition of the GPT given in \eqref{eq:PigG}, one gets:
$$P\{\emptyset\}=0$$
$$P\{\theta_1\}=m(\theta_1)+\frac{1}{2}m(\theta_2)+m(\theta_1\cap\theta_2)+\frac{2}{3}m(\theta_1\cup\theta_2)$$
$$P\{\theta_2\}=m(\theta_2)+\frac{1}{2}m(\theta_1)+m(\theta_1\cap\theta_2)+\frac{2}{3}m(\theta_1\cup\theta_2)$$
\begin{equation*}
\begin{split}
P\{\theta_1\cap\theta_2\}=\frac{1}{2}m(\theta_2)+\frac{1}{2}m(\theta_1)+\\
m(\theta_1\cap\theta_2)+\frac{1}{3}m(\theta_1\cup\theta_2)
\end{split}
\end{equation*}
\begin{equation*}
\begin{split}
P\{\theta_1\cup\theta_2\} = P\{\Theta\} = m(\theta_1)+ m(\theta_2) + \\
m(\theta_1\cap\theta_2) + m(\theta_1\cup\theta_2)=1
\end{split}
\end{equation*}
\noindent It is easy to prove that $0\leq P\{.\} \leq1$ and $P\{\theta_1\cup\theta_2\}=P\{\theta_1 \}+P\{ \theta_2\} -P\{\theta_1\cap\theta_2\}$
\item
{\bf{With the Shafer's model}}: If one adopts the Shafer's model (we assume $\theta_1\cap\theta_2\overset{\mathcal{M}^0}{\equiv}\emptyset$), then after applying the hybrid DSm rule of combination, one gets a basic belief assignment with non null masses only on $\theta_1$, $\theta_2$ and $\theta_1\cup\theta_2$. By applying the GPT, one gets:
$$P\{\emptyset\}=0 \qquad P\{\theta_1\cap\theta_2\}=0$$
$$P\{\theta_1\}=m(\theta_1)+ \frac{1}{2}m(\theta_1\cup\theta_2)$$
$$P\{\theta_2\}=m(\theta_2)+ \frac{1}{2}m(\theta_1\cup\theta_2)$$
$$P\{\theta_1\cup\theta_2\}=m(\theta_1)+ m(\theta_2)+m(\theta_1\cup\theta_2)=1$$ 
\noindent which naturally corresponds in this case to the pignistic probability built with the classical pignistic transformation \eqref{eq:Pig}.
\end{itemize}

%*******************************************
\subsection{Example for the 3D case}
%*******************************************

\begin{itemize}
\item
{\bf{With the free DSm model}}: Let's  consider $\Theta=\{\theta_1,\theta_2,\theta_3\}$, its hyper-power set $D^\Theta=\{\alpha_0,\ldots,\alpha_{18}\}$ (with $\alpha_i$, $i=0,\ldots, 18$ corresponding to propositions explicated in table 1 of section 2.4), and the generalized basic belief assignment $m(.)$ over the hyper-power set $D^\Theta$. The six tables presented in the appendix show the full derivations of all pignistic probabilities $P\{\alpha_i\}$ for $i=1,\ldots,18$ ($P\{\emptyset\}$ always equals zero) according to the GPT formula \eqref{eq:PigG}.\\

Note that $P\{\alpha_{18}\}=1$ because $(\theta_1\cup\theta_2\cup\theta_3)$ corresponds to the sure event in our subjective probability space and $\sum_{\alpha_i\in D^\Theta}m(\alpha_i)=1$ by the definition of any generalized basic belief assignment $m(.)$ defined on $D^\Theta$.\\

It can be verified (as expected) on this example, although being a quite tedious task, that the Poincar\'{e}' equality holds:
\begin{equation*}     
P\{A_{1}\cup\ldots\cup A_{n}\}=\underset{I\neq\emptyset}{\sum_{I\subset      \{1,\ldots,n\}}}      {(-1)}^{\vert I \vert+1}	     P\{\bigcap_{i\in I} A_{i}\}	 	
\end{equation*}
It is also easy to verify that $\forall A\subset B \Rightarrow P\{A\} \leq P\{B\}$ holds. 
By example, for $(\alpha_6\triangleq(\theta_1\cup\theta_3)\cap\theta_2) \subset \alpha_{10}\triangleq\theta_2)$ and from the expressions of $P\{\alpha_6\}$ and $P\{\alpha_{10}\}$ given in appendix, we directly conclude that $P\{\alpha_6\}\leq P\{\alpha_{10}\}$ because for all $X \in D^\Theta$, $\frac{\mathcal{C}_{\mathcal{M}}(X\cap \alpha_6)}{\mathcal{C}_{\mathcal{M}}(X)}\leq \frac{\mathcal{C}_{\mathcal{M}}(X\cap \alpha_{10})}{\mathcal{C}_{\mathcal{M}}(X)}$ as shown in the following table
\begin{equation*}
\begin{array}{|c|ccc|}
\hline
X & \frac{\mathcal{C}_{\mathcal{M}}(X\cap \alpha_6)}{\mathcal{C}_{\mathcal{M}}(X)} & \leq & \frac{\mathcal{C}_{\mathcal{M}}(X\cap \alpha_{10})}{\mathcal{C}_{\mathcal{M}}(X)} \\
\hline 
\alpha_1 & 1       & \leq & 1  \\
\alpha_2 & 1       & \leq & 1   \\ 
\alpha_3 & (1/2) & \leq &  (1/2)  \\ 
\alpha_4 & 1 & \leq &  1    \\ 
\alpha_5 & (2/3) & \leq & (2/3)\\
\alpha_6 & 1 & \leq &  1  \\ 
\alpha_7 & (2/3) & \leq &  (2/3)   \\ 
\alpha_8 & (3/4) & \leq & (3/4)  \\
\alpha_9 &  (2/4) & \leq &   (2/4) \\ 
\alpha_{10} &  (3/4) & \leq & 1 \\
\alpha_{11} &   (2/4) & \leq &   (2/4)   \\
\alpha_{12} &    (3/5) & \leq & (3/5)  \\ 
\alpha_{13} &    (3/5) & \leq & (4/5) \\ 
\alpha_{14} &    (3/5) & \leq & (3/5) \\ 
\alpha_{15} &    (3/6) & \leq & (4/6) \\ 
\alpha_{16} &    (3/6) & \leq &  (3/6) \\ 
\alpha_{17} &    (3/6) & \leq & (4/6) \\ 
\alpha_{18} &    (3/7) & \leq & (4/7)\\
\hline
    \end{array}
\end{equation*}

\item
{\bf{Example with a given hybrid DSm model}}: Consider now the hybrid model $\mathcal{M}\neq\mathcal{M}^f$ in which we force all possible conjunctions to be empty, but $\theta_1\cap\theta_2$ according to the second Venn diagram presented in section 2.4. In this case the hyper-power set $D^\Theta$ reduces to 9 elements $\{\alpha_0,\ldots,\alpha_8\}$ explicated in  table 2 of section 2.4.The following tables present the full derivations of the pignistic probabilities $P\{\alpha_i\}$ for $i=1,\ldots,8$ from the GPT formula \eqref{eq:PigG} applying to this hybrid model.
\begin{equation*}
\begin{array}{|l|l|l|}
\hline
P\{\alpha_1\}= & P\{\alpha_2\}= & P\{\alpha_3\}=\\
 (1/1)m(\alpha_1) &  (0/1)m(\alpha_1) &  (1/1)m(\alpha_1) \\
 +(0/1)m(\alpha_2) & +(1/1)m(\alpha_2) & +(0/2)m(\alpha_2)\\
 +(1/2)m(\alpha_3) & +(0/2)m(\alpha_3)  & +(2/2)m(\alpha_3)\\
+(1/2)m(\alpha_4) & +(0/2)m(\alpha_4)  & +(1/2)m(\alpha_4)\\
 +(1/3)m(\alpha_5) & +(0/3)m(\alpha_5)  & +(2/3)m(\alpha_5)\\
 +(1/3)m(\alpha_6) & +(1/3)m(\alpha_6)  & +(2/3)m(\alpha_6)\\
+(1/3)m(\alpha_7) & +(1/3)m(\alpha_7)  & +(1/3)m(\alpha_7)\\
+(1/4)m(\alpha_8) & +(1/4)m(\alpha_8)  & +(2/4)m(\alpha_8)\\
\hline
\end{array}
\end{equation*}
\begin{center}
{\bf{Table 4}}: Derivation of $P\{\alpha_1\triangleq\theta_1\cap\theta_2\}$, $P\{\alpha_2\triangleq\theta_3\}$ and $P\{\alpha_3\triangleq\theta_1\}$
\end{center}
\begin{equation*}
\begin{array}{|l|l|l|}
\hline
P\{\alpha_4\}= & P\{\alpha_5\}= & P\{\alpha_6\}=\\
 (1/1)m(\alpha_1) &  (1/1)m(\alpha_1) &  (1/1)m(\alpha_1) \\
 +(0/1)m(\alpha_2) & +(0/1)m(\alpha_2) & +(1/1)m(\alpha_2)\\
 +(1/2)m(\alpha_3) & +(2/2)m(\alpha_3)  & +(2/2)m(\alpha_3)\\
+(2/2)m(\alpha_4) & +(2/2)m(\alpha_4)  & +(1/2)m(\alpha_4)\\
 +(2/3)m(\alpha_5) & +(3/3)m(\alpha_5)  & +(2/3)m(\alpha_5)\\
 +(1/3)m(\alpha_6) & +(2/3)m(\alpha_6)  & +(3/3)m(\alpha_6)\\
+(2/3)m(\alpha_7) & +(2/3)m(\alpha_7)  & +(2/3)m(\alpha_7)\\
+(2/4)m(\alpha_8) & +(3/4)m(\alpha_8)  & +(3/4)m(\alpha_8)\\
\hline
\end{array}
\end{equation*}
\begin{center}
{\bf{Table 5}}: Derivation of $P\{\alpha_4\triangleq\theta_2\}$, $P\{\alpha_5\triangleq\theta_1\cup\theta_2\}$ and $P\{\alpha_6\triangleq\theta_1\cup\theta_3\}$
\end{center}
\begin{equation*}
\begin{array}{|l|l|}
\hline
P\{\alpha_7\}= & P\{\alpha_8\}=\\
  (1/1)m(\alpha_1) &  (1/1)m(\alpha_1) \\
 +(2/2)m(\alpha_2) & +(2/2)m(\alpha_2) \\
 +(1/2)m(\alpha_3) & +(2/2)m(\alpha_3)  \\
+(2/2)m(\alpha_4) & +(2/2)m(\alpha_4)  \\
 +(2/3)m(\alpha_5) & +(3/3)m(\alpha_5)  \\
 +(2/3)m(\alpha_6) & +(3/3)m(\alpha_6)  \\
+(3/3)m(\alpha_7) & +(3/3)m(\alpha_7)  \\
+(3/4)m(\alpha_8) & +(4/4)m(\alpha_8)  \\
\hline
\end{array}
\end{equation*}
\begin{center}
{\bf{Table 6}}: Derivation of $P\{\alpha_7\triangleq\theta_2\cup\theta_3\}$ and $P\{\alpha_8\triangleq\theta_1\cup\theta_2\cup\theta_3\}$
\end{center}

\item
{\bf{Example with the Shafer's model}}: Consider now the Shafer's model $\mathcal{M}^0\neq\mathcal{M}^f$ in which we force all possible conjunctions to be empty according to the third Venn diagram presented in section 2.4. In this case the hyper-power set $D^\Theta$ reduces to the classical power set $2^\Theta$ with 8 elements $\{\alpha_0,\ldots,\alpha_7\}$ explicated in  table 3 of section 2.4. Applying, the GPT formula \eqref{eq:PigG}, one gets the following pignistic probabilities $P\{\alpha_i\}$ for $i=1,\ldots,7$ which naturally coincide, in this particular case, with the values obtained directly by the classical pignistic transformation \eqref{eq:Pig}:
\begin{equation*}
\begin{array}{|l|l|l|}
\hline
P\{\alpha_1\}= & P\{\alpha_2\}= & P\{\alpha_3\}=\\
 (1/1)m(\alpha_1) &  (0/1)m(\alpha_1) &  (0/1)m(\alpha_1) \\
 +(0/1)m(\alpha_2) & +(1/1)m(\alpha_2) & +(0/1)m(\alpha_2)\\
 +(0/1)m(\alpha_3) & +(0/1)m(\alpha_3)  & +(1/1)m(\alpha_3)\\
+(1/2)m(\alpha_4) & +(1/2)m(\alpha_4)  & +(0/2)m(\alpha_4)\\
 +(1/2)m(\alpha_5) & +(0/2)m(\alpha_5)  & +(1/2)m(\alpha_5)\\
 +(0/2)m(\alpha_6) & +(1/2)m(\alpha_6)  & +(1/2)m(\alpha_6)\\
+(1/3)m(\alpha_7) & +(1/3)m(\alpha_7)  & +(1/3)m(\alpha_7)\\
\hline
\end{array}
\end{equation*}
\begin{center}
{\bf{Table 7}}: Derivation of $P\{\alpha_1\triangleq\theta_1\}$, $P\{\alpha_2\triangleq\theta_2\}$ and $P\{\alpha_3\triangleq\theta_3\}$
\end{center}
\begin{equation*}
\begin{array}{|l|l|l|}
\hline
P\{\alpha_4\}= & P\{\alpha_5\}= & P\{\alpha_6\}=\\
 (1/1)m(\alpha_1) &  (1/1)m(\alpha_1)    &  (0/1)m(\alpha_1) \\
 +(1/1)m(\alpha_2) & +(0/1)m(\alpha_2) & +(1/1)m(\alpha_2)\\
 +(0/1)m(\alpha_3) & +(1/1)m(\alpha_3)  & +(1/1)m(\alpha_3)\\
+(2/2)m(\alpha_4) & +(1/2)m(\alpha_4)  & +(1/2)m(\alpha_4)\\
 +(1/2)m(\alpha_5) & +(2/2)m(\alpha_5)  & +(1/2)m(\alpha_5)\\
 +(1/2)m(\alpha_6) & +(1/2)m(\alpha_6)  & +(2/2)m(\alpha_6)\\
+(2/3)m(\alpha_7) & +(2/3)m(\alpha_7)  & +(2/3)m(\alpha_7)\\
\hline
\end{array}
\end{equation*}
\begin{center}
{\bf{Table 8}}: Derivation of $P\{\alpha_4\triangleq\theta_1\cup\theta_2\}$, $P\{\alpha_5\triangleq\theta_1\cup\theta_3\}$ and $P\{\alpha_6\triangleq\theta_2\cup\theta_3\}$
\end{center}
\begin{equation*}
\begin{array}{|l|}
\hline
P\{\alpha_7\}= \\
 (1/1)m(\alpha_1) \\
 +(1/1)m(\alpha_2) \\
 +(1/1)m(\alpha_3) \\
+(2/2)m(\alpha_4) \\
 +(2/2)m(\alpha_5) \\
 +(2/2)m(\alpha_6) \\
+(3/3)m(\alpha_7) \\
\hline
\end{array}
\end{equation*}
\begin{center}
{\bf{Table 9}}: Derivation of $P\{\alpha_7\triangleq\theta_1\cup\theta_2\cup\theta_3\}=1$
\end{center}
\end{itemize}

\section{ Conclusion}
%--------------------------------------------------------------------------------------------

A generalization of the classical pignistic transformation developed originally within the DST framework has been proposed in this work. This generalization is based on the new theory of plausible and paradoxical reasoning (DSmT) and provides a new mathematical issue to help the decision-making under uncertainty and  paradoxist (i.e. highly conflicting) sources of information. The generalized pignistic transformation (GPT) proposed here allows to build a subjective/pignistic probability measure over the hyper-power set of the frame of the problem under consideration 
or any kind of model (free, hybrid or Shafer's model). 
The GPT coincides naturally with the classical pignistic transformation whenever the Shafer's model is adopted. It corresponds with assumptions of classical pignistic probability generalized to the free DSm model. A relation of GPT on general hybrid models to assumptions of classical PT is still in the process of investigation.
Several examples for the 2D and 3D cases for different kinds of models have been presented to illustrate the validity of the GPT.
%
%for any kind of model (free, hybrid or Shafer's model). The GPT coincides naturally with the classical pignistic transformation whenever the Shafer's is adopted. Several examples for the 2D and 3D cases for different kinds of models have been presented to illustrate the validity of the GPT.

%%%%%%%%%%%%%%%%%%%%%%%%%%%%%%%%%%
% Reference list

%\bibliographystyle{unsrt}
%{\small
%\bibliography{Fusion_2004_IFA_v1}
%}

\clearpage
\newpage
\onecolumn
\section*{Appendix: Derivation of the GPT for the 3D free DSm model}

\begin{tabular}{cc}
   \includegraphics[height=7cm]{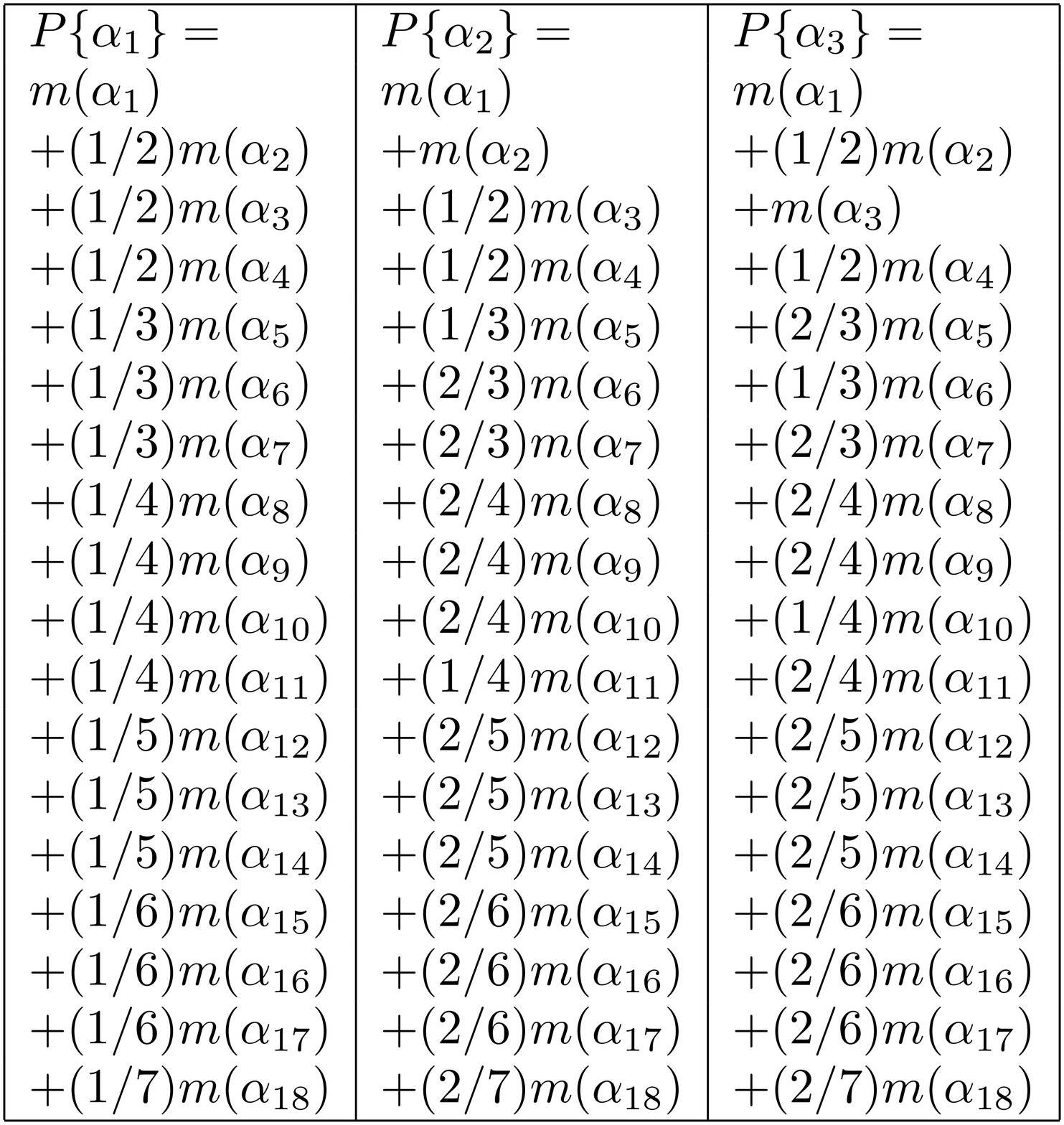} &\includegraphics[height=7cm]{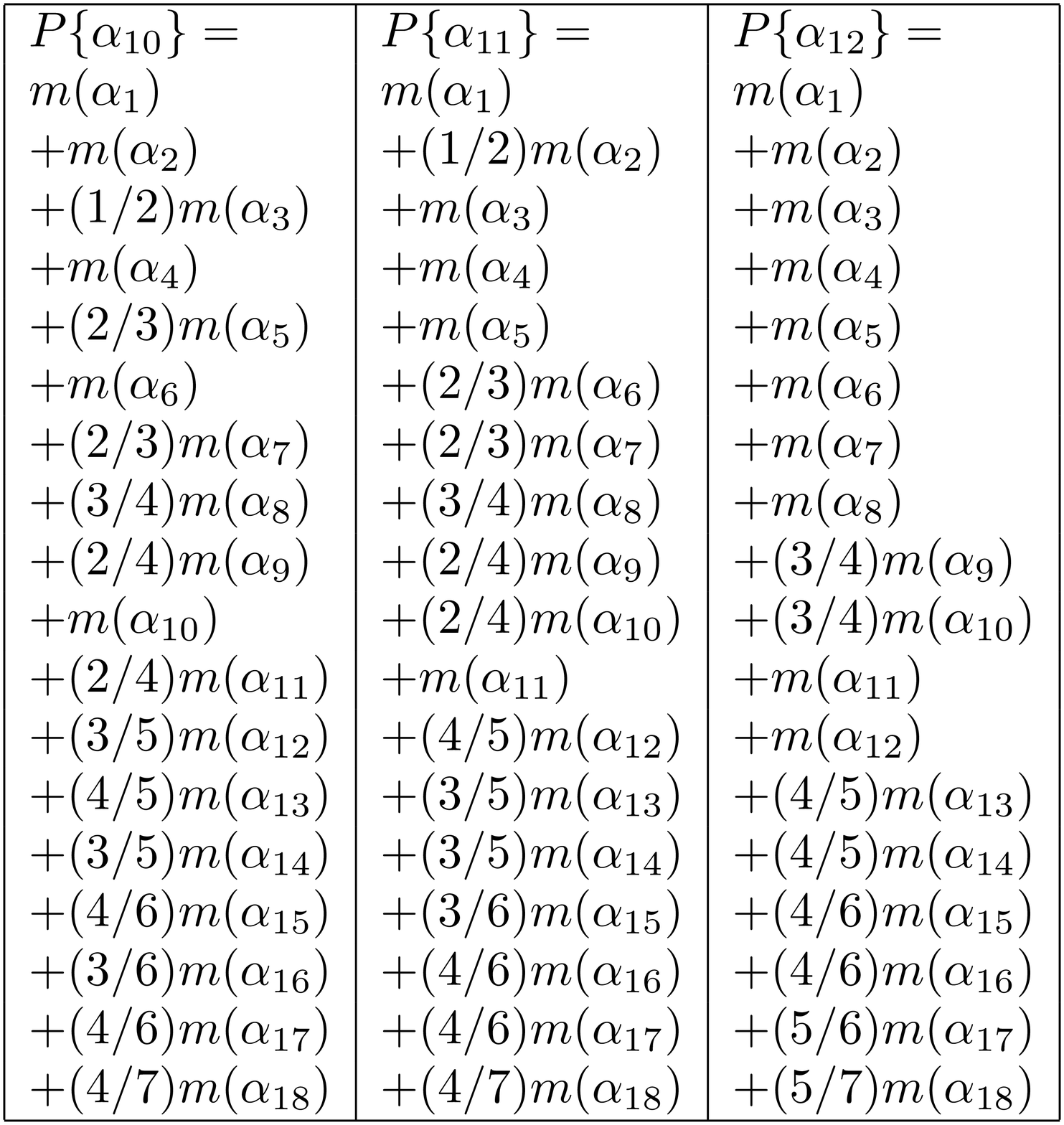} \\
{\bf{Table 10}}: Derivation of $P\{\alpha_1\}$, $P\{\alpha_2\}$ and $P\{\alpha_3\}$ & {\bf{Table 13}}: Derivation of $P\{\alpha_{10}\}$, $P\{\alpha_{11}\}$ and $P\{\alpha_{12}\}$
\end{tabular}
\vfill
\begin{tabular}{cc}
   \includegraphics[height=7cm]{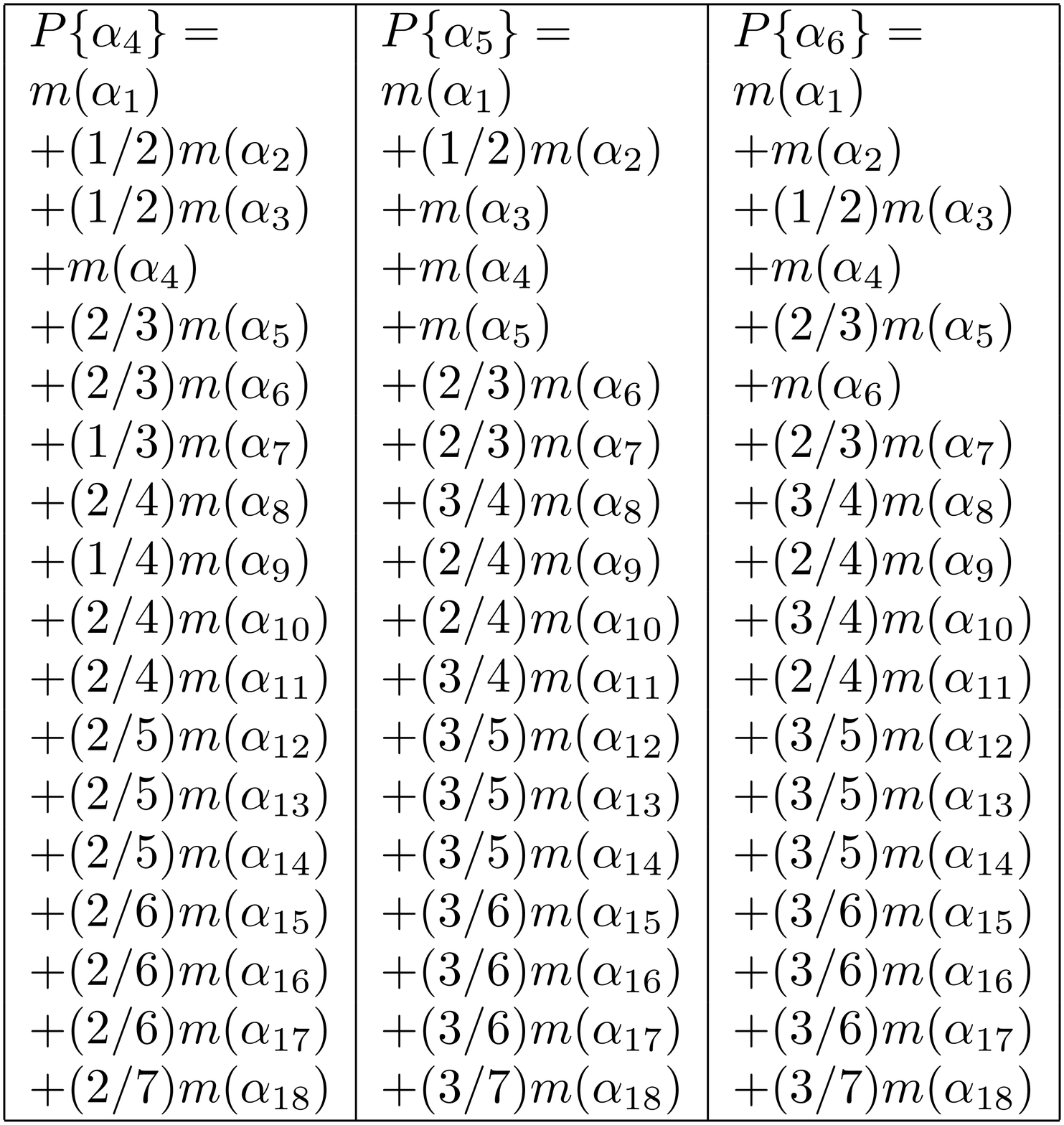} &\includegraphics[height=7cm]{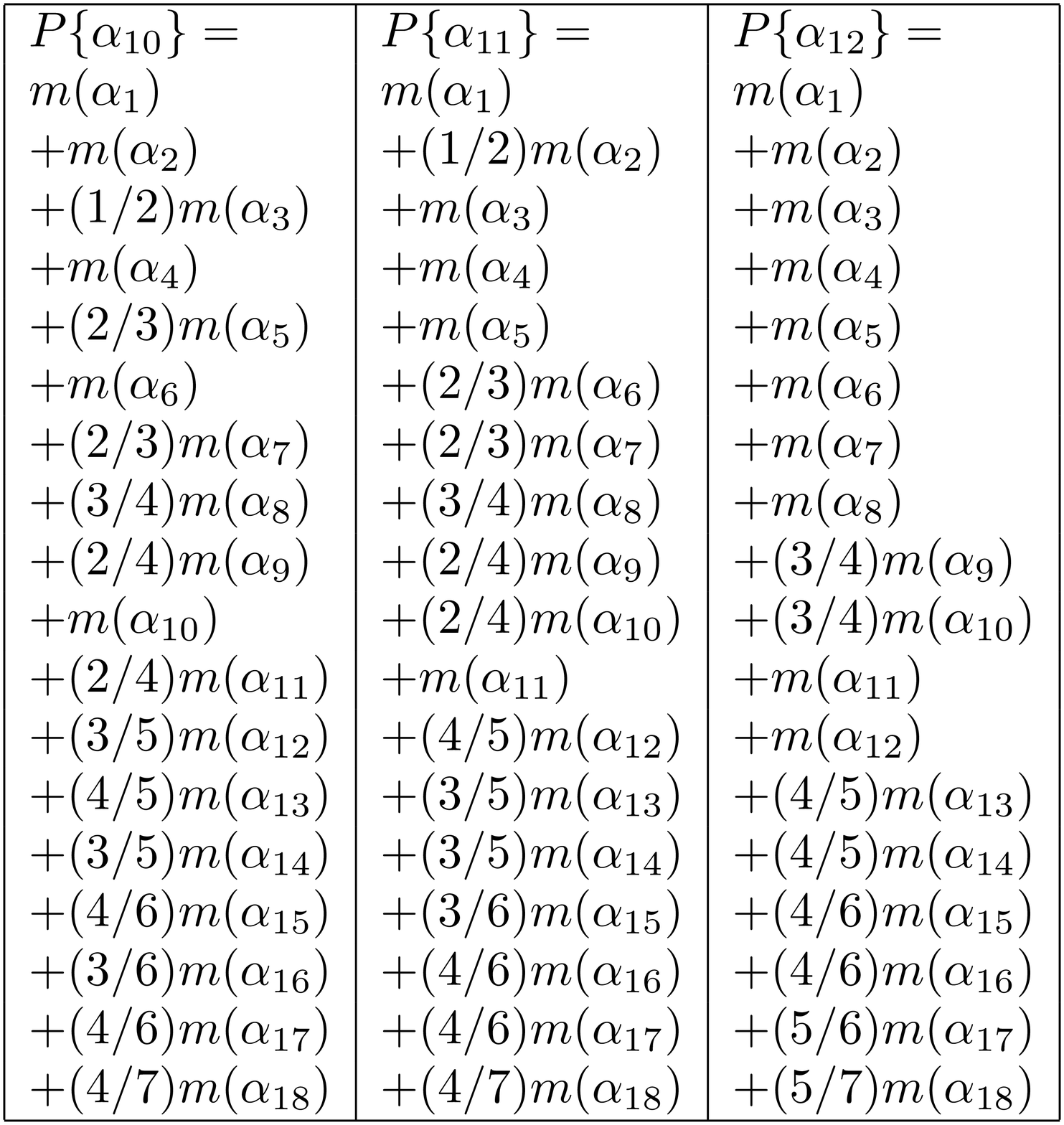} \\
{\bf{Table 11}}: Derivation of $P\{\alpha_4\}$, $P\{\alpha_5\}$ and $P\{\alpha_6\}$ & {\bf{Table 14}}: Derivation of $P\{\alpha_{13}\}$, $P\{\alpha_{14}\}$ and $P\{\alpha_{15}\}$
\end{tabular}
\vfill
\begin{tabular}{cc}
   \includegraphics[height=7cm]{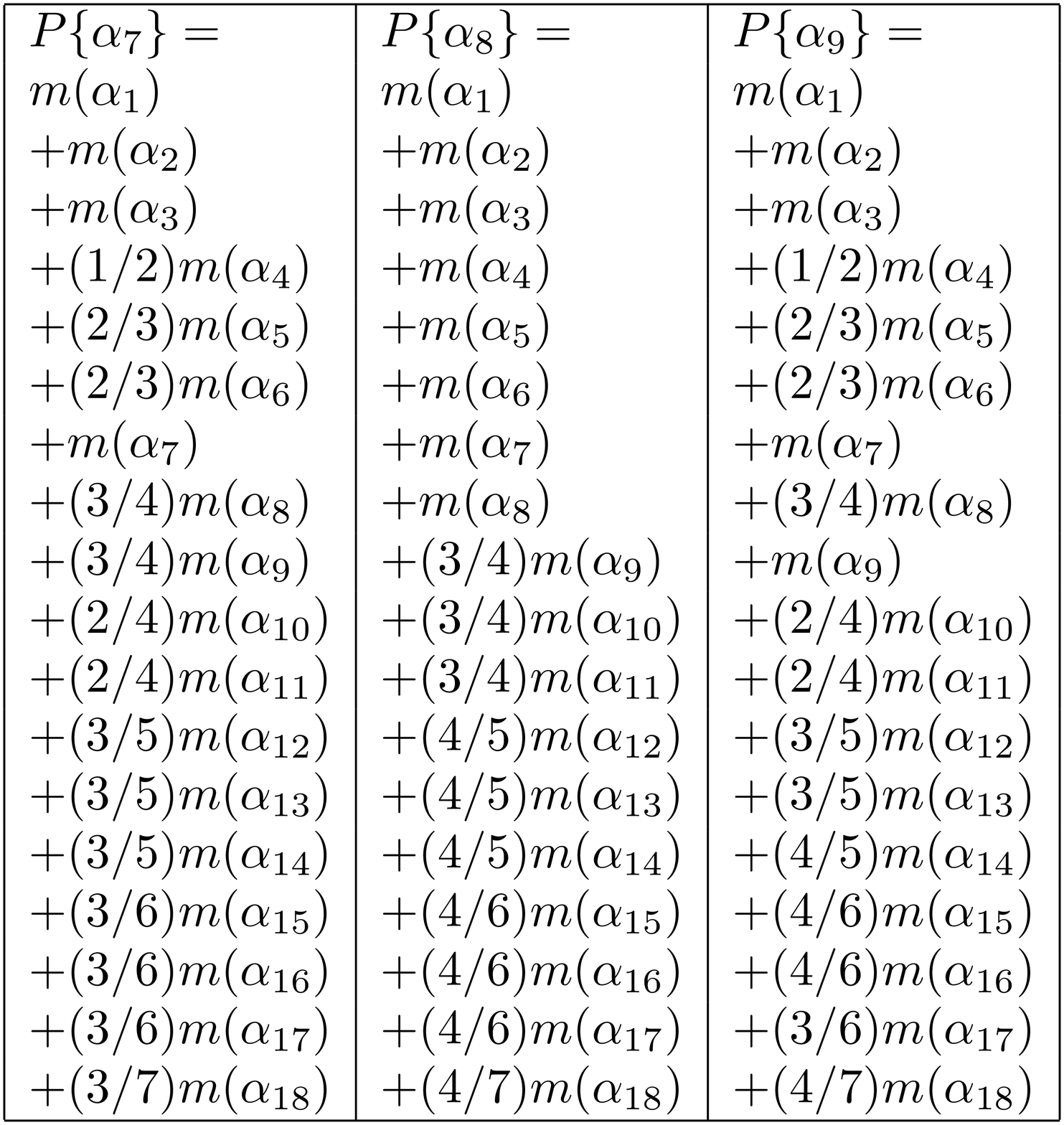} &\includegraphics[height=7cm]{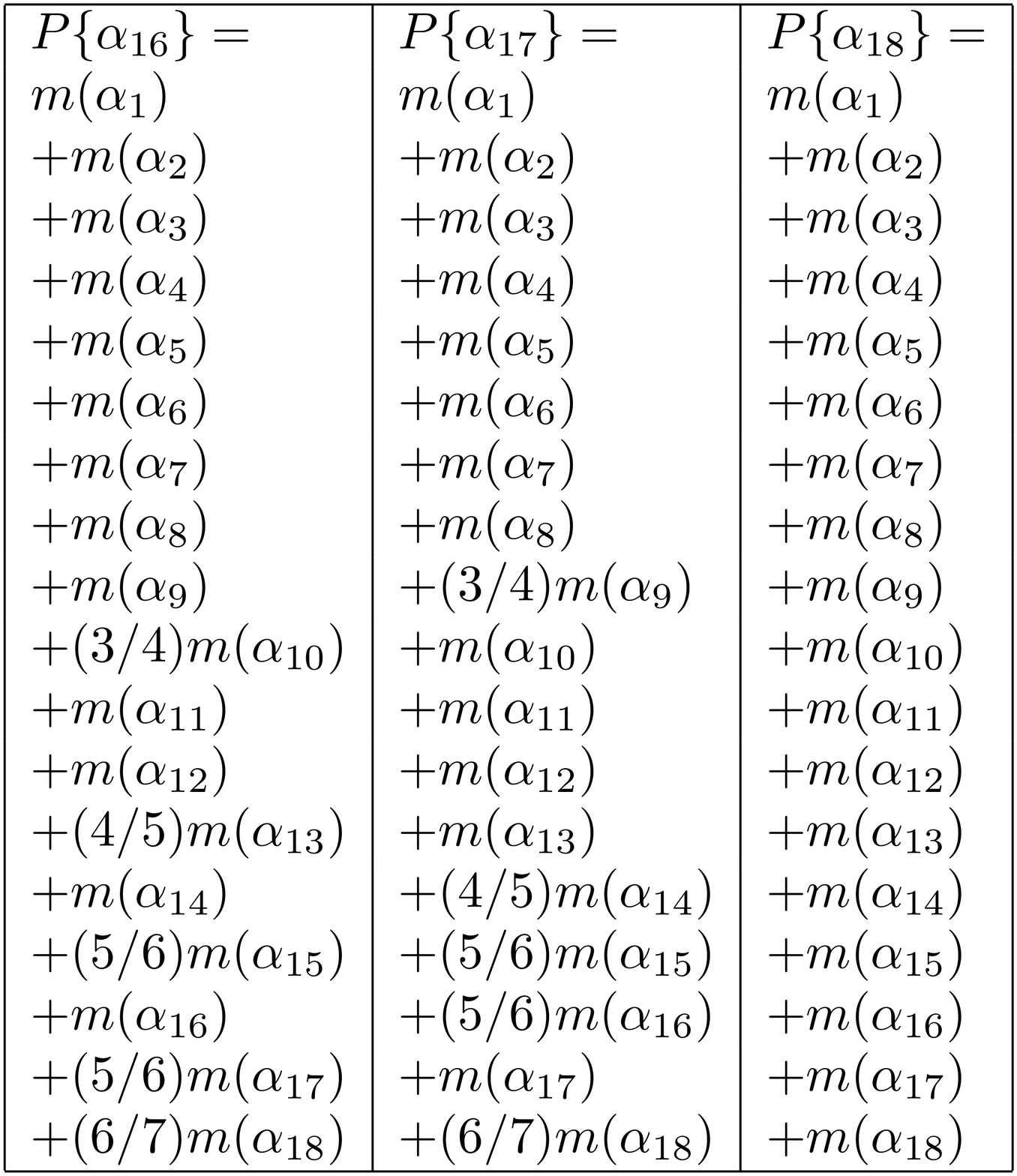} \\
{\bf{Table 12}}: Derivation of $P\{\alpha_7\}$, $P\{\alpha_8\}$ and $P\{\alpha_9\}$ & {\bf{Table 15}}: Derivation of $P\{\alpha_{16}\}$, $P\{\alpha_{17}\}$ and $P\{\alpha_{18}\}$
\end{tabular}

\end{document}